\documentclass{article}
\usepackage{arxiv}

\usepackage[utf8]{inputenc} % allow utf-8 input
\usepackage[T1]{fontenc}    % use 8-bit T1 fonts
\usepackage{hyperref}       % hyperlinks
\usepackage{url}            % simple URL typesetting
\usepackage{booktabs}       % professional-quality tables
\usepackage{nicefrac}       % compact symbols for 1/2, etc.
\usepackage{microtype}      % microtypography
\usepackage{lipsum}
\usepackage{commands}
\usepackage{multirow}
\usepackage{tikz-cd}
\usepackage{mathtools}
\usepackage{float}
\usepackage{afterpage}
\usepackage{tabularx}
\usepackage{adjustbox}

\usepackage{cite}
\usepackage{amsmath,amssymb,amsfonts}  % blackboard math symbols
\usepackage{algorithmic}
\usepackage{graphicx}
\usepackage{textcomp}
\usepackage{xcolor}
\usepackage{scalerel}
\usepackage{tikz}
\usetikzlibrary{svg.path}

\definecolor{orcidlogocol}{HTML}{A6CE39}
\tikzset{
  orcidlogo/.pic={
    \fill[orcidlogocol] svg{M256,128c0,70.7-57.3,128-128,128C57.3,256,0,198.7,0,128C0,57.3,57.3,0,128,0C198.7,0,256,57.3,256,128z};
    \fill[white] svg{M86.3,186.2H70.9V79.1h15.4v48.4V186.2z}
                 svg{M108.9,79.1h41.6c39.6,0,57,28.3,57,53.6c0,27.5-21.5,53.6-56.8,53.6h-41.8V79.1z M124.3,172.4h24.5c34.9,0,42.9-26.5,42.9-39.7c0-21.5-13.7-39.7-43.7-39.7h-23.7V172.4z}
                 svg{M88.7,56.8c0,5.5-4.5,10.1-10.1,10.1c-5.6,0-10.1-4.6-10.1-10.1c0-5.6,4.5-10.1,10.1-10.1C84.2,46.7,88.7,51.3,88.7,56.8z};
  }
}

\newcommand\orcidicon[1]{\href{https://orcid.org/#1}{\mbox{\scalerel*{
\begin{tikzpicture}[yscale=-1,transform shape]
\pic{orcidlogo};
\end{tikzpicture}
}{|}}}}

\def\BibTeX{{\rm B\kern-.05em{\sc i\kern-.025em b}\kern-.08em
    T\kern-.1667em\lower.7ex\hbox{E}\kern-.125emX}}

\usepackage{commands}

\begin{document}
% \title{SegFlow: A Conditional Normalizing Flow Model for the Data Augmentation of Samples in Synthetic Seismic Images Datasets}
\title{Generating Data Augmentation samples for Semantic Segmentation of Salt Bodies in a Synthetic Seismic Image Dataset}


\author{
    Luis~Felipe~Henriques~\orcidicon{0000-0002-0608-0625} \\
      Department of Computer Science\\
      Pontifical Catholic University of Rio de Janeiro\\
      Rio de Janeiro, Brazil \\
      \texttt{lhenriques@inf.puc-rio.br} \\
  \And
    Sérgio~Colcher~\orcidicon{0000-0002-3476-8718} \\
      Department of Computer Science\\
      Pontifical Catholic University of Rio de Janeiro\\
      Rio de Janeiro, Brazil \\
      \texttt{colcher@inf.puc-rio.br} \\
  \And
    Ruy~Luiz~Milidiú~\orcidicon{0000-0002-3423-9998} \\
      Department of Computer Science\\
      Pontifical Catholic University of Rio de Janeiro\\
      Rio de Janeiro, Brazil \\
      \texttt{milidiu@inf.puc-rio.br} \\
  \And
    André~Bulcão~\orcidicon{0000-0002-9871-9683} \\
        CENPES\\
        % Centro de Pesquisa e Desenvolvimento Leopoldo A. Miguez de Mello\\
        PETROBRAS\\
        Rio de Janeiro, Brazil\\ 
        \texttt{bulcao@petrobras.com.br} \\
  \And
    Pablo~Barros~\orcidicon{0000-0002-8835-6276}\\
        CENPES\\
        % Centro de Pesquisa e Desenvolvimento Leopoldo A. Miguez de Mello\\
        PETROBRAS\\
        Rio de Janeiro, Brazil\\ 
        \texttt{pablobarros@petrobras.com.br}
}

\maketitle

\begin{abstract}
% \begin{abstract}
    Nowadays, 
        subsurface salt body localization and delineation, 
        also called semantic segmentation of salt bodies, 
        are among the most challenging geophysicist tasks. 
    Thus, 
        identifying large salt bodies is notoriously tricky and is crucial for identifying hydrocarbon reservoirs and drill path planning. 
    While several successful attempts to apply Deep Neural Networks (DNNs) have been made in the field, 
        the need for a huge amount of labeled data and the associated costs of manual annotations by experts sometimes prevent the applicability of these methods. 
    This work proposes a Data Augmentation method based on training two generative models to augment the number of samples in a seismic image dataset for the semantic segmentation of salt bodies.
    Our method uses deep learning models to generate pairs of seismic image patches and their respective salt masks for the Data Augmentation.
    % Our method consists of training two distinct deep learning models to generate pairs of seismic image patches and their respective salt masks and using them as data augmentations. 
    The first model is a Variational Autoencoder and is responsible for generating patches of salt body masks. 
    The second is a Conditional Normalizing Flow model, 
        which receives the generated masks as inputs and generates the associated seismic image patches. 
    We evaluate the proposed method by comparing the performance of ten distinct state-of-the-art models for semantic segmentation, 
        trained with and without the generated augmentations, 
        in a dataset from two synthetic seismic images. 
    The proposed methodology yields an average improvement of $8.57\%$ in the IoU metric across all compared models. 
    The best result is achieved by a DeeplabV3+ model variant, 
        which presents an IoU score of $95.17\%$ when trained with our augmentations. 
    Additionally, 
        our proposal outperformed six selected data augmentation methods, 
        and the most significant improvement in the comparison, of $9.77\%$, is achieved by composing our DA with augmentations from an elastic transformation.
    At last, 
        we show that the proposed method is adaptable for a larger context size by achieving results comparable to the obtained on the smaller context size.

\end{abstract}

% sections area
\section{Introduction} \label{sec:1-introduction}
    
    Seismic imaging methods are used to produce images of the Earth’s subsurface properties from seismic data.
    The seismic data, 
        also called seismogram, 
        is recorded on the surface by geophone devices 
        that capture elastic waves emitted by artificial sources 
        and reflected in the subsoil.
    % The seismic data, 
    %     also called seismogram, 
    %     is recorded on the surface by geophone devices that capture elastic waves emitted by artificial sources that are reflected in the subsoil.
    Further, 
        seismograms are processed to form images representing the boundaries of the different rock types in the Earth’s subsurface \cite{SamuelHGray2016}.
    
    Nowadays, 
        seismic images are applied to a broad range of seismic applications: 
        from near-surface environmental studies 
        to oil and gas exploration 
        and even to long-period earthquake seismology.
    Among different applications in the industry, 
        seismic imaging analysis plays a paramount role 
        in the exploration 
        and identification of hydrocarbon fuel reservoirs.
    The task of 
        salt deposit localization and delineation 
        is crucial in such application since overlying rock-salt formations can trap hydrocarbons reservoirs due to their exceeding impermeability \cite{Babakhin2019SemiSupervisedSO}.
    Unfortunately, 
        the exact identification of large salt bodies is notoriously tricky \cite{Jones2014SeismicII} 
        and often requires manual interpretation of seismic images by domain experts.

    Several automation tools addressing the manual interpretation requirements have been proposed 
        \cite{Bedi2018SFAGTMSF, Di2018MultiattributeKC, Hegazy2014TextureAF, Pitas1992ATA,Wrona2018SEISMICFA, Wu2016MethodsTC, Zhao2015ACO}. 
    However, 
        these methods do not generalize well for complex cases since they rely on handcrafted features.
    Consequently, 
        many works emerged proposing Deep Neural Networks (DNNs) for the semantic segmentation of salt bodies since they are a natural way to overcome the need for hand-engineering features.
    Although those DNNs present superior performance on the task compared to traditional methods \cite{Dramsch2018DeeplearningSF, Waldeland2018ConvolutionalNN, Wang2018AutomaticSD, Zeng2019AutomaticSS}, 
        their requirement of huge amounts of labeled data and the associated costs of expert manual annotation prevents their wider applicability.

    From the machine learning perspective, 
        the semantic segmentation of salt bodies 
        consists of predicting binary salt masks that depict salt regions 
        from the inputted seismic image patches.

    This work aims to mitigate the requirement of a huge amount of manually labeled data 
        when training DNNs for the semantic segmentation of salt bodies.
    With such a goal, 
        we propose a Data Augmentation (DA) method to increase the number of samples in a dataset for the semantic segmentation of salt bodies. 
    Our method consists of training two distinct deep learning models 
        to generate pairs of seismic image patches 
        and salt masks for using them as DA.

    The first model is a Variational Autoencoder \cite{Kingma2013AutoEncodingVB} 
        whose is responsible for generating salt body masks. 
    The second is a Conditional Normalizing Flow model \cite{Dinh2014NICENI, Dinh2016DensityEU, Kingma2018GlowGF, Ardizzone2019GuidedIG},
        which receives the generated masks as inputs 
        and outputs the associated seismic image patches.
    In each generated pair, 
        the mask is responsible for depicting (and bounding) the salt body, 
        and the generated seismic image patch is a seismic representation of the respective salt body.
    We conduct experiments on a dataset from two synthetic seismic images and their velocity models, 
        from where the masks are extracted by thresholding and clipping. 
    The images are taken from distinct and publicly available synthetic seismic models \cite{Stoughton20012DEM, Aminzadeh19953DMP} designed to emulate salt prospects in deep water,
        such as those found in the Gulf of Mexico. 
    We compare state-of-the-art semantic segmentation models on this dataset 
        when augmenting and not augmenting 
        the data with the generated samples. 
    The comparison is performed for a plain U-net model \cite{Ronneberger2015UNetCN} 
        and nine DeeplabV3+ \cite{deeplabv3plus2018} variants. 
        % with 
        % the Xception \cite{Chollet2016XceptionDL}, 
        % MobileNet \cite{mobilenetv32019}, 
        % and Resnet \cite{He2016DeepRL} backbones.

    Our main contribution is developing a DA method for the semantic segmentation of salt bodies that effectively improves the quality of machine learning models regardless of their architectures and sizes. 
    We also show that the proposed method outperforms several DA methods when applied standalone.
        % , usually applied for semantic segmentation models. 
    At last, 
        we show that the proposed DA method is easily adapted to a different dataset setup, 
        such as a larger context size, 
        without losing efficiency. 

    %%%%%%%%%%%%%%%%%%%%%%%%%%%%
        
    The remainder of this paper is organized as follows.
    In section \ref{sec:2-related-work}, 
        we review the most relevant related work on subsurface salt body localization and delineation.
    In section \ref{sec:3-Background}, 
        we offer a background section, 
        where we give a brief introduction to Conditional Normalizing Flows 
        and Variational Autoencoders models.
    Next, 
        section \ref{sec:4-model} describes our proposed methodology and detailed models’ architectures.
    Section \ref{sec:5-dataset} provides a short dataset description.
    The experiments 
        and results are shown in section \ref{sec:6-experiments}. 
    Finally, 
        we close the paper with a brief conclusion in section \ref{sec:7-conclusion}.

\section{Related Work} \label{sec:2-related-work}
        The semantic segmentation of salt deposits in a seismic image is a problem 
        that has been attracted many researchers over the years. 
    The traditional approach is based on handcrafting different feature extractors before computing the appropriate responses. 
    
    In 1992, 
        Pitas and Kotropoulos \cite{Pitas1992ATA} presented a method for the semantic segmentation of seismic images based on texture analyses and other handcrafted features.
        % suggested a method based on texture analyses for semantic segmentation of seismic images.
    Since from, many methods using texture-based attributes have been proposed.
    Hegazy and Ghassan \cite{Hegazy2014TextureAF} 
        presented in 2014 a technique that combines the three texture attributes along with region boundary smoothing for delineating salt boundaries.
    In 2015, 
        Shafiq, M. et al. \cite{Shafiq2015DetectionOS} proposed using the gradient of textures as a seismic attribute claiming that it can quantify texture variations in seismic volumes.
    A year later, 
        they engineered a saliency-based feature to detect salt dome bodies within seismic volumes \cite{Shafiq2016SalSiAN}.   
    
    Several methods address the task by detecting the salt deposit boundaries.
    In 2015, 
        Amin and Deriche \cite{Asjad2015ANA} proposed an approach based on using a 3D multi-directional edge detector. 
    Wu (2016) \cite{Wu2016MethodsTC} relies on a likelihood estimation method for the salt boundaries identification. 
    In 2018, 
        Di et al. \cite{Di2018MultiattributeKC} presented an unsupervised workflow for delineating the surface of salt bodies from seismic surveying based on a multi-attribute k-means cluster analysis.

    The advent of deep neural networks (DNNs) 
        and their impressive results in the semantic segmentation of natural and medical images
        \cite{Ronneberger2015UNetCN, deeplabv3plus2018, Cheng2020PanopticDeepLabAS, Zhu2019ImprovingSS} 
        motivated several works to the semantic segmentation of seismic images. 
    In 2018, 
        Dramsch and Lüthje \cite{Dramsch2018DeeplearningSF}  evaluated several classification DNNs with transfer learning to identify nine different seismic textures from $65 \times 65$ pixel patches.
    Di et al. \cite{Di2018RealtimeSI} proposed a deconvolutional neural network (DCNN) for supporting real-time seismic image interpretation. 
    They evaluated their proposal performance in an application 
        for segmenting the F3 seismic dataset \cite{F3Dataset}.
    In a second work \cite{Di2018DeepCN}, 
         they addressed the salt body delineation using Convolutional Neural Networks (CNNs) in $32 \times 32$ image patches from a synthetic dataset.
     Using the same approach,
        Waldeland et al. \cite{Waldeland2018ConvolutionalNN} presented a custom CNN for the semantic segmentation of salt bodies in 3D seismic data split into cubes of $65 \times 65 \times 65$.
    
    Already in 2019, 
        Zeng et al. \cite{Zeng2019AutomaticSS} 
        showed the great potential and benefits of applying CNNs for salt-related interpretations by using a state-of-art network with a U-Net structure along with the residual learning framework \cite{He2016IdentityMI, He2016DeepRL}. 
    They achieved high precision in the salt body delineation task in a stratified K-fold cross-validation setting on the SEG-SEAM data \cite{SEGSEAM}.
    Their work also explored network adjustments, 
        including the Exponential Linear Units (ELU) activation function \cite{Clevert2016FastAA} 
        and the Lovasz-Softmax loss function \cite{Rakhlin2018LandCC}. 
    Shi et al. \cite{Shi2019SaltSegA3}
        formulated the problem as 3D image segmentation 
        and presented an efficient approach based on CNNs with an encoder-decoder architecture. 
    They trained the model by randomly extracting sub-volumes from a large-scale 3D dataset
        to feed into the network. 
    They also applied data augmentation.
    Wu et al. \cite{Wu2019FaultSeg3DUS} introduced a very similar approach, 
        which used CNNs to 3D salt segmentation in a dataset composed of 
        $200$ 3D synthetic seismic images split into 
        $128 \times 128 \times 128$ sub-volume patches.
    They also used the class-balanced binary cross-entropy loss function to, 
        during the training procedure, 
        address the imbalance between salt and non-salt areas present in their dataset.
    
    Still, in 2019,
        the TGS Salt Identification Challenge
        \footnote{https://www.kaggle.com/c/tgs-salt-identification-challenge/} 
        was released.
    Babakhin et al. \cite{Babakhin2019SemiSupervisedSO} 
        describe the first-place solution in their paper. 
    They propose a semi-supervised method 
        for the segmentation of salt bodies in seismic images, 
        which utilizes unlabeled data for multi-round self-training. 
    To reduce error amplification in the self-training, 
        they propose a scheme that uses an ensemble of CNNs. 
    They achieved the state-of-the-art on the TGS Salt Identification Challenge dataset.
    
    Finally, 
        in 2020 Milosavljevic \cite{Milosavljevic2020IdentificationOS} presents a method based on training a U-Net architecture combined with the ResNet and the DenseNet\cite{Huang2017DenselyCC} networks.
    To better comprehend the properties of the proposed architecture, 
        they present an ablation study on the network components and a grid search analysis on the size of the applied ensemble.
    
    The class imbalance between salt and non-salt areas,
        the huge amount of labeled data needed to train models, 
        the focus on delineating salt bodies, 
        strategies for the data augmentation, 
        and unlabeled data usage are commonly discussed in previous work.  
    Such statements are the inspiration for this work. 
    We propose a DA method for semantic segmentation of salt bodies models.
    % We propose a methodology for the data augmentation of the training data for DNN models for the semantic segmentation of salt bodies.
    The proposal is to train two DNN models 
        that generate pairs of seismic image patches 
        and their respective salt masks 
        containing at least $10\%$ 
        and a maximum of $90\%$ salt on each patch.
    Thus, 
        using the generated pairs for data augmentation indirectly gives more importance (or weight) for samples in the salt deposit boundary areas.
        % of the entire seismic image. 
    This work is the first attempting such an approach for the semantic segmentation of salt bodies to the best of our knowledge.

    % Such statements are the inspiration for this work,
    %     where we propose a methodology for the data augmentation of the training data for salt body segmentation models. 
    % The proposal is to build a machine learning system composed of two models that generate pairs of seismic image patches 
    %     and their respective salt masks containing at least $20\%$ and a maximum of $80\%$ salt on each patch. 
    % Thus, 
    %     using the generated pairs for data augmentation indirectly gives more importance (or weight) for examples in the salt deposit boundary areas of the entire seismic image.
    % This work is the first attempting such an approach for the semantic segmentation of salt bodies to the best of our knowledge.

\section{Background} \label{sec:3-Background}
    
    This section discusses recent efforts in the Variational Autoencoders (VAEs) and Conditional Invertible Normalizing Flows (CNF).
    % This section discusses recent efforts in Variational Autoencoders (VAEs); 
    % Invertible Normalizing Flows(NF);
    % and its conditional extension -- the Invertible Conditional Normalizing Flows (CNF).

    \subsection{Variational Autoencoder} \label{sec:vae}
    
    Consider a probabilistic model $p_{\theta}(x, z)$ with observations $x$,
        continuous latent variables $z$, 
        and model parameters $\theta$. 
    In generative modeling, 
        we are often interested in learning the model parameters $\theta$ by estimating the marginal likelihood $p_{\theta}(x)$. 
    Unfortunately, 
        the marginalization over the unobserved latent variables $z$ is generally intractable \cite{Kingma2013AutoEncodingVB}.
    
    Instead, 
        variational inference \cite{Jordan1999AnIT} constructs a lower bound on the marginal likelihood logarithm $log\ p_{\theta}(x)$. 
    The lower bound is built by introducing a variational approximation to the posterior.
    
        \begin{equation}
            \label{eq:elbo}
            log\ p_{\theta}(x) \geq \mathbb{E}_{q} \left [\ log\ p_{\theta}(x|z)\ \right ] - \mathbb{KL}(q_{\phi}(z|x) || \pi(z)) 
        \end{equation}
    
    Equation \ref{eq:elbo} is called as 
        the Evidence Lower Bound (ELBO), 
        where $q_{\phi}(z|x)$ is the posterior approximation, 
        $p_{\theta}(x|z)$ is a likelihood function, 
        $\pi(z)$ is the prior distribution, 
        and $\phi$ and $\theta$ are the model parameters.
    Thus, 
        VAEs \cite{Kingma2013AutoEncodingVB} are generative models 
        that use neural networks 
        to predict the variational distribution parameters.
    
    From the machine learning perspective,
        VAEs are defined by:
        an encoder model which maps $x$ to $z$;
        a decoder model which maps $z$ back to $x$;
        and a loss function that is the negative of equation \ref{eq:elbo}.
        % defines the loss function. 
        % and is expressed by two terms. 
    The loss first term is a likelihood function 
        that is minimized during the model training. 
    The second term is the Kullback-Leibler Divergence between 
        the inferred posterior distribution $q_{\phi}(z|x)$ 
        and the prior distribution $\pi(z)$. 
    It can be interpreted as a regularization term 
        that enforces the distribution of the latent variables.  
    Furthermore, 
        the $\pi(z)$ distribution is a standard and diagonal Gaussian. 
    It is worth mentioning that the better the posterior approximation, 
        the tighter the ELBO is, 
        and so smaller is the gap between the true distribution $p(x)$
        and the lower bound. 
    The whole model is trained through stochastic maximization of ELBO through the reparameterization trick \cite{Kingma2013AutoEncodingVB, Rezende2014StochasticBA}.
    
    In this work, we propose using a VAE to estimate the distribution of salt masks presented in our dataset and thus learning a stochastic generation process of salt masks.

    % The model parameters
    %     $\theta$ and $\phi$ are jointly trained through stochastic maximization of ELBO through the reparameterization trick \cite{Kingma2013AutoEncodingVB, Rezende2014StochasticBA}.

    % It is worth mentioning that the better the posterior approximation, 
    %     the tighter the ELBO is, 
    %     and so smaller is the gap between the true distribution $p(x)$ and the lower bound.
    % However, 
    %     a diagonal-covariance Gaussian is usually chosen as the variational distribution $q_{\phi}(z|x)$.
    % With such a simple distribution, 
    %     the ELBO will be fairly loose,
    %     resulting in biased maximum likelihood estimates \cite{Berg2018SylvesterNF}. %*of the model parameters  
    % Therefore,  
    %     a more flexible posterior distribution is needed.
    \subsection{Conditional Normalizing Flow Models} \label{sec:normflow}
    
    Normalizing Flow (NF) models 
        \cite{Dinh2014NICENI, Dinh2016DensityEU, Kingma2018GlowGF} 
        learn flexible distributions by transforming a simple base distribution 
        with a sequence of invertible transformations, 
        known as normalizing flows. 
    The Conditional Normalizing Flow (CNF) 
        \cite{Ardizzone2019GuidedIG, Winkler2019LearningLW, Lugmayr2020SRFlowLT} 
        extends the NF by conditioning the entire generative process on the input data, 
        commonly referred to as conditioning data.

    This work aims to estimate conditional likelihoods 
        of the seismic image patches $x$, 
        given their respective salt masks $y$, 
        which is the conditioning data. 
    Thus, 
        given three vector-spaces $x \in R^{d}$, 
        $x \in R^{d}$, 
        and $z \in R^{d}$, 
        an invertible mapping $f_{\omega}: x \rightarrow z$ 
        with $\omega$ parameters, 
        and a base distribution $\pi(z_{k}|y)$ 
        then CNF compute the data log-likelihood $log\ p(x|y)$ 
        using the change of variables formula:

            \begin{equation}
                \label{eq:change-of-variable-formula}
                log\ p(x|y) = log\ \pi(f_{\omega}(x)|y) + 
                    log \left | 
                        det 
                        \left (
                                \frac
                                    { \partial f_{\omega}(x) }
                                    {\partial x} 
                            \right ) 
                        \right |
            \end{equation}
    
    In equation \ref{eq:change-of-variable-formula}, 
        the term $| det( \partial f_{\omega}(x)/ \partial x ) |$ 
        is the Jacobian Determinant absolute value, 
        or for simplicity, 
        the Jacobian of $f_{\omega}$. 
    It measures the change in density 
        when going from $x$ to $z$ 
        under the transformation $f_{\omega}$. 
    In general, 
        the cost of computing the Jacobian will be $O(D^{3})$. 
    However, 
        it is possible to design transformations in which the Jacobians are efficiently calculated \cite{Kingma2018GlowGF}.
    
    Thus, 
        given an input $y$ 
        and a target $x$, 
        CNF estimates the distribution $p(x|y)$ using the conditional base distribution $\pi(z_{k}|y)$ 
        and a sequence of $k$ mappings 
        $
            z_{k} = f_{\omega}(x) = f^{k}(...(f^{1}(x)))
        $
        which are bijective in $x$.
    
    The main challenge in invertible normalizing flows 
        is to design mappings to compose the transformation $f_{\omega}$ 
        since they must have a set of restricting properties 
        \cite{Winkler2019LearningLW}. 
    All mappings $f^{i} \in f_{\omega}$ must have 
        a known and tractable inverse mapping 
        and an efficiently computable Jacobian 
        while being powerful enough to model complex transformations. 
    Moreover, 
        fast sampling is desired, so the inverse mappings $(f^{i})^{-1}$ should be calculated efficiently.
    
    Fortunately, 
        the current literature on normalizing flows provides a set of well-established invertible layers with all those properties. 
    % Fortunately, 
    %     there is a set of well-established invertible layers with all those properties in the current literature of normalizing flows. 
    The CNF model proposed in this work is built upon step-flow blocks \cite{Kingma2018GlowGF}, 
        which is a stacking of three invertible layers
        \footnote{Due to space constraints we refer to \cite{Kingma2018GlowGF} for more details on those invertible layers.}: 
            (I) the Actnorm layer \cite{Kingma2018GlowGF}; 
            (II) the Invertible 1 × 1 Convolution layer \cite{Kingma2018GlowGF}; 
            (III) the Affine Coupling layer \cite{Dinh2014NICENI, Dinh2016DensityEU}.
            
    Additionally, 
        the step-flow blocks are combined with the multi-scale architecture \cite{Dinh2016DensityEU} 
        that allows the model to learn intermediate representation at different scale levels that are fine-grained local features 
        \cite{Rezende2014StochasticBA, Salakhutdinov2009DeepBM, Dinh2016DensityEU}.
    At regular intervals, 
        step-flow blocks are surrounded by 
        squeezing and factor-out operations, 
        which reduces the spatial resolution 
        and doubles the channel dimension, 
        producing multiple scale levels. 
    % Thus, 
    %     at regular intervals, 
    %     step-flow blocks are surrounded by squeezing and factor-out operations,
    %     producing multiple levels of scale. 
    % The spatial resolution is reduced at each scale, 
    %     and the number of hidden layer features is doubled.
    This approach reduces the computation and memory used by the model 
        and distributes the loss function throughout the network. 
    It is similar to guiding intermediate layers using intermediate classifiers \cite{Lee2014DeeplySupervisedN}.
    
    In such a model, 
        a major component is the Affine Coupling layer which is expressed as:
            \begin{equation}
                \label{eq:affine-coupling-layer}
                h =
                \left \{
                \begin{matrix}
                    h_{1:d} = x_{1:d}; \\ 
                    h_{d+1:D} = x_{d+1:D} \odot  e^{s(x_{1:d})} + t(x_{1:d})
                \end{matrix}
                \right.
            \end{equation}
    In equation \ref{eq:affine-coupling-layer}, 
        $\odot $ is the element-wise product and, 
        $s(x_{1:d})$ and $t(x_{1:d})$ are the outputs of the layer’s backbone,
        a standard neural network block. 
    The input is split into two halves $x = [x_{1:d}; x_{d+1:D}]$, 
        and the split $x_{d+1:D}$ is updated using the $s$ and  $t$ operators computed by the layer’s backbone from the $x_{1:d}$ split. 
    The layer’s output is the concatenation of the haves 
        $h = [h_{1:d}; h_{d+1:D}]$.
    
    In summary, 
        an observation $x$ is given as inputs to the model 
        during the training procedure 
        and passed through the invertible mapping $f_{\omega}(x)$ 
        producing $z_{k}$. 
    Then, 
        the base distribution parameters and the likelihood are evaluated using the conditioning data $y$ such that:
            \begin{equation}
                \label{eq:prior-net}
                \pi(z_{k}|y) = \mathcal{N}(z; \mu(y), \sigma(y)),
            \end{equation}
        where $\mu(y)$ and $\sigma(y)$ 
        are the outputs of a simple convolutional neural network that uses the conditioning term $y$ as input.
    In the remainder, 
        this convolutional network is referred to as Prior Network.
    % As usual, 
    %     in neural networks, 
    %     the flow and base distribution parameters are optimized using stochastic gradient descent.
    % The conditional data $y$ is introduced to the base distribution such that:
    %     \begin{equation}
    %         \label{eq:prior-net}
    %         \pi(z_{k}|y) = \mathcal{N}(z; \mu(y), \sigma(y)),
    %     \end{equation}
    % where $\mu(y)$ and $\sigma(y)$ 
    % are the outputs of a simple convolutional neural network that uses the conditioning term $y$ as input. 
    % In the remainder, 
    %     this convolutional network is referred to as Prior Network.
    
    Finally, 
        the generative process is done by sampling $z_{k}$ 
        from the base density $\pi(z_{k}|y)$ 
        and then transformed by the sequence of inverse mappings 
        $x = f_{\omega}^{-1}(z_{k})$.
        % which has its parameters determined by $y$ 
        % and the Prior Network. 
        % that receives y as input. 
    % Then, 
    %     $z_{k}$ is transformed by the sequence of inverse mappings 
    %     $f_{\omega}^{-1}(z_{k})$. 
    % Therefore, 
    %     the generative process is defined as:
    %     \begin{equation}
    %         \label{eq:z-follows-p}
    %         % z_{k} \sim \pi(f_{\omega}(x,y)|y)
    %         z_{k} \sim \pi(f_{\omega}(x)|y)
    %     \end{equation}
    %     \begin{equation}
    %         \label{eq:x-fz}
    %         % x = f_{\omega}^{-1}(z_{k},y)
    %         x = f_{\omega}^{-1}(z_{k})
    %     \end{equation}

    % Equation \ref{eq:affine-coupling-layer-inverse} defines the inverse transformation.
    %     \begin{equation}
    %         \label{eq:affine-coupling-layer-inverse}
    %         x = 
    %         \left\{
    %         \begin{matrix}
    %             x_{1:d} = h_{1:d}; \\ 
    %             x_{d+1:D} = (h_{d+1:D} - t(h_{1:d})) \odot  exp(-s(h_{1:d}))
    %         \end{matrix}
    %         \right.
    %     \end{equation}

    % In the following, 
    %     we present the invertible layers that we use in this work.
    
    % \input{content/background/invertible_layers/coupling-layers}
    % \input{content/background/invertible_layers/invertible-conv}
    % \input{content/background/invertible_layers/actnorm}
    % TODO: stepflow block

    % \input{sections/3-Background/conditional_flow}
\section{Methodology} \label{sec:4-model}
    
    This section describes the proposed method for generating the data augmentation samples 
        and the related models.

    Our approach is learning to generate sample pairs of seismic image patches
        and their respective salt masks 
        that lie in the boundary of a salt body. 
    The primary motivation for this approach is that 
        if we can correctly segment all salt body boundaries, 
        it is straightforward to recover all entire bodies -- 
        since everything inside a boundary is salt. 
    Also, 
        since the most common strategy to train models 
        for the salt semantic segmentation is 
        to split seismic images into patches 
        and then using those as training examples, 
        patches lying in the boundary of a salt body usually have low frequency in the datasets. 
    As a consequence, 
        learning to segment such examples becomes a challenge for the models.
    Therefore, 
        our methodology consists of training two generative models to generate pairs of seismic image patches 
        and their respective salt masks.

    The first model is a standard VAE model used for sampling salt body masks from the dataset’s inferred distribution. 
    Only the dataset masks $y$ are given as input to the model during the training, 
        while a lower bound (the ELBO), 
        defined by equation \ref{eq:elbo}, 
        on the mask distributions is optimized. 
    Such as discussed in section \ref{sec:3-Background}, 
        at inference time, 
        we sample the latent variables $z$ from the prior distribution 
        $\pi(z) \sim \mathbb{N}(0,\mathbb{I})$ 
        and then decode $z$ with the VAE’s decoder to generate a salt mask $\hat{y}$ belonging to the inferred distribution.
    
    The second model is a Conditional Normalizing Flow model, 
        which receives the sampled salt body masks $\hat{y}$ as inputs to generate the associated seismic image patches $\hat{x}$. 
    During the training, 
        the CNF model receives the seismic image patches $x$ 
        and their associated salt masks $y$ as inputs. 
    Then the log-likelihood of the patches $log\ p(x|y)$ is maximized using the change of variable formula, 
        equation \ref{eq:change-of-variable-formula}. 
    The generative process is done by sampling $z_{k}$ 
        from the base distribution 
        $\pi(z_{k}) = \mathbb{N}(\mu(\hat{y}), \sigma(\hat{y}))$, 
        which has its parameters determined by the Prior Network.
        % ,
        % a simple CNN that receives the sampled salt masks $y$ as input. 
    At last, 
        $z_{k}$ is transformed by the CNF’s a sequence of bijective mappings, resulting in the sampled image patch 
        $\hat{x} = (f_{\omega}^{-1}(z_{k})$.
    
    It is worth emphasizing that both models, 
        the VAE and CNF, 
        are trained separately 
        and independently of each other 
        and are only used together to generate the samples for the data augmentation. 
    Figure 1 illustrates the generation procedure in a flow chart. 
        \begin{figure}[H]
            \centering
            \includegraphics[width=0.6\linewidth]
            {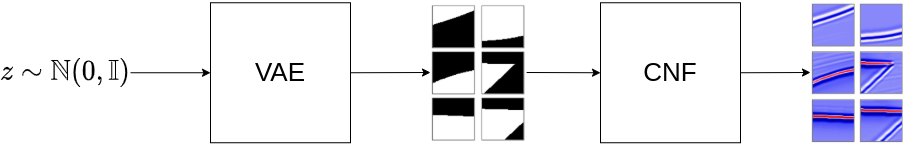}
            \caption{Data augmentation generation procedure.}
            \label{fig:generation-procedure}
        \end{figure}
    The procedure is simple.
    First, 
        we use the trained VAE model to sample salt masks $\hat{y}$,
        and then the image patches are sampled using $\hat{y}$ as conditional data.
    
    In the following, 
        we describe the setup, 
        hyper-parameter settings, 
        and learning procedure for the VAE and CNF models trained in this work.
    
    \subsection*{Variational AutoEncoder}
\begin{figure}[H]
        \centering
        \includegraphics[width=0.5\linewidth]
        {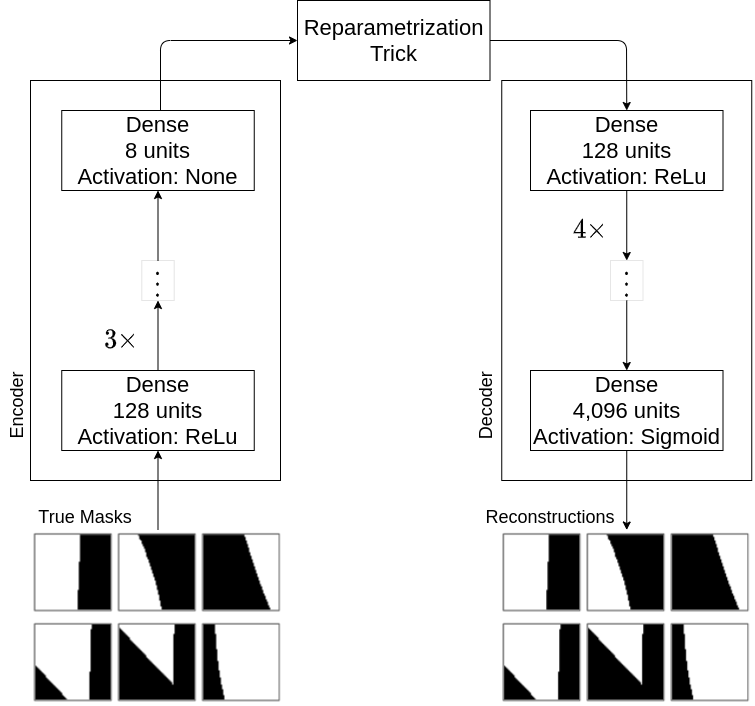}
        \caption{VAE layers composition.}
        \label{fig:vae-diagram}
    \end{figure}

    We train a simple VAE \cite{Kingma2013AutoEncodingVB} model used to sample salt masks 
        given to the CNF model as conditional data during the data augmentation generation procedure. 
    The VAE encoder and decoder 
        are just simple feed-forward networks. 
    The encoder contains four stacked dense layers. 
        % where the first three layers have $128$ units 
        % and use the ReLu activation function. 
    % The last layer has eight units, 
    %     and no activation function is applied.
    In its turn, 
        the decoder contains five stacked dense layers. 
        % in which the first four layers have $128$ units 
        % and uses the ReLu activation function. 
    % The last layer has $4,096$ units 
        % and uses the sigmoid activation function. 
    Figure \ref{fig:vae-diagram} illustrates the VAE architecture.
    Moreover,
        the model inputs are flattened, 
        and the outputs are reshaped to form images.
        
    Finally, 
        the VAE model is trained via gradient descent with mini-batches of size $300$ to minimize the loss described by equation \ref{eq:elbo}. 
    Thus, 
        the model is trained for $65.6K$ iterations using the Adam optimizer \cite{Kingma2014AdamAM} with its default parameters, 
        and the initial learning rate is $10^{-3}$. 
    Additionally, 
        the learning rate is scheduled to decay in a polynomial form, with a warm-up phase of $6.5K$ steps. 
    The training procedure takes about $8$ hours and $13$ minutes on an NVidia Tesla P100 GPU, 
        and the training loss converged around $31.54$.

% loss : 31.54146957397461
% epoch: 434
% time per epoch: 35
    \subsection*{Conditional Normalizing Flows}
    We train a CNF model, 
        which, 
        during the data augmentation generation procedure, 
        receives the sampled masks $\hat{y}$ as inputs 
        to generate the associated seismic image patches $\hat{x}$.
    
\begin{figure}[H]
        \centering
        \includegraphics[width=0.7\linewidth]
        {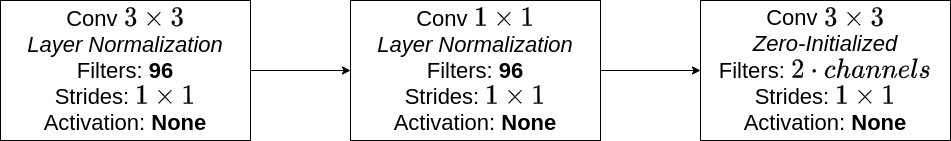}
        \caption{Coupling Layer's backbone}
        \label{fig:s-and-t-block}
    \end{figure}

    The CNF uses multi-scale architecture, 
        with four scale levels and $15$ step-flow blocks per level.
    The Invertible $1 \times 1$ Convolution outputs are given as inputs 
        to the Affine Coupling Layer at each block, which has its backbone network illustrated in figure \ref{fig:s-and-t-block}. 
    % Subsequently, 
    %     the inputs are split into two halves. 
    % The layer's backbone calculates the $s$ and $t$ operators using the input's first half. 
    % This backbone network is a standard neural network block 
    %     and is illustrated in Figure \ref{fig:s-and-t-block}.

    All backbone weights are initialized by the Xavier Normal initialization method \cite{Glorot2010UnderstandingTD}, 
        except for the weights of the last convolutional layer, 
        which is initialized with zeros.
    
\begin{figure}[ht!]
            \centering
            % \begin{tikzcd}
            %     \text{Conv $3 \times 3$} \arrow{r} &
            %     \text{ReLu} \arrow{r} &
            %     \text{$\cdots$} \arrow{r} &
            %     \text{Conv $3 \times 3$} \arrow{r} &
            %     \text{ReLu} \arrow{r} &
            %     \text{Conv $1 \times 1$}
            % \end{tikzcd}
            \includegraphics[width=0.5\linewidth]
            {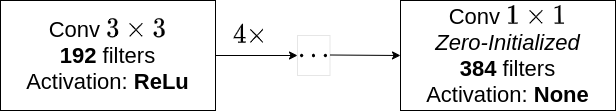}
            \caption{Prior Network.}
            \label{fig:prior-net}
        \end{figure}

    In it turns, 
        the Prior Network is composed of 4 stacked $3 \times 3$ convolutions followed by a last zero-initialized $1 \times 1$ convolutional layer.
        % with the ReLu as the activation function followed by a last zero-initialized $1 \times 1$ convolutional layer. 
    Figure \ref{fig:prior-net} illustrates the prior network. 
    Additionally, 
        padding and strides are set such that each $3 \times 3$ convolutional layer halves the inputs' spatial dimension and
        % , 
        % and $192$ filters are applied. 
        the last $1 \times 1$ convolutional layer doubles the $z$ channel dimensions 
        with $384$ filters. 
        % while maintaining its spatial dimension.
    
    The model is trained to minimize the input data negative log-likelihood,
        calculated through equation \ref{eq:change-of-variable-formula}. 
    Moreover, 
        during the training procedure, 
        we regularize the model with the Binary Cross-Entropy loss to ensure spatial correlation between the learned latent space variables $z$ 
        and the conditional data $y$. 
    Such a regularization is analogous to the regularization proposed in the Glow paper \cite{Kingma2018GlowGF}, 
        which penalizes the miss-classifications of a linear layer when learning a class-conditioned model.
    
\begin{figure}[ht!]
            \centering
            % \begin{tikzcd}
            %     \text{Transposed Conv $3 \times 3$} \arrow{r} &
            %     \text{$\cdots$} \arrow{r} &
            %     \text{Transposed Conv$3 \times 3$} \arrow{r} &
            %     \text{Conv $3 \times 3$}
            % \end{tikzcd}
            \includegraphics[width=0.6\linewidth]
            {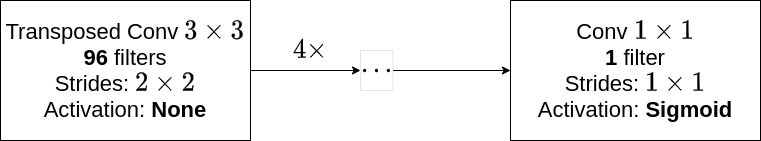}
            \caption{Auxiliary network, used to compute the regularization term during the training procedure.}
            \label{fig:aux-network}
        \end{figure}

    To compute the Binary Cross-Entropy loss, 
        we use an auxiliary network jointly trained with the whole CNF model.
    Such as in the Conditional Glow, 
        this auxiliary network receives the latent variable $z$ as input 
        and outputs the predicted respective mask. 
    In summary, 
        the auxiliary network comprises four stacked $3 \times 3$ transposed convolutions \cite{Zeiler2010DeconvolutionalN}, followed by a $3 \times 3$ convolutional layer
        where each transposed convolution doubles the input's spatial resolution. 
        % and by the sigmoid activation function. 
    Figure \ref{fig:aux-network} illustrates this auxiliary network.
    
    % Each transposed convolution doubles the input's spatial resolution 
    %     and has $96$ filters. 
    % In it turn, 
    %     the convolutional layer applies a single filter to its inputs, 
    %     and the sigmoid activation ensures that the output values are between zero and one.

    Finally, 
        the CNF model is trained on the training and validation sets, 
        using the Adam Optimizer \cite{Kingma2014AdamAM}, 
        with its default hyper-parameters, during $396.8K$ iterations 
        and mini-batch size $50$. 
    Polynomial decay is applied to the learning rate, 
    which goes from $10^-4$ to $0$.
    % The learning rate is scheduled to decay in a polynomial form from $1 \times 10^-4$ to $0$. 
    The training procedure takes about four days on a Google Cloud TPU v2-8,
    and the training loss converges around $1.67$.
\section{Dataset} \label{sec:5-dataset}
    We conduct extensive experiments on a dataset 
        from two publicly available synthetic salt prospects models, 
        and our results are presented on that dataset. 
    The Pluto1.5 dataset \cite{Stoughton20012DEM}, 
        and the SEG/EAGE Salt model \cite{Aminzadeh19953DMP}, 
        are designed to emulate salt prospects in deep water, 
        such as those found in the Gulf of Mexico.
    
    We extract a 2D migrated image 
        from each seismic model, 
        and the binary salt mask is extracted from its velocity model by thresholding 
        and clipping. 
    Additionally, 
        each migrated image is normalized by its mean and standard deviation.
    
\begin{figure}[ht!]
        \centering
        \includegraphics[keepaspectratio=true,width=0.95\columnwidth]
        {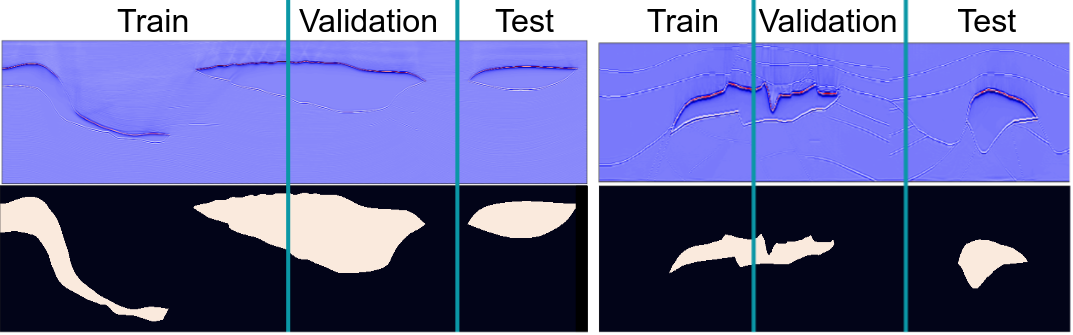}
        \caption{
            Seismic migrated images and their binary masks. 
            The first row in the figure presents two entire seismic images, 
            and the second row shows their corresponding salt masks. 
            The black regions in the salt masks represent non-salt areas in their respective seismic images. 
            The vertical bars in the image delimitate the locations of each image 
            that are designated for the dataset's train, validation, and test sets.
        }
        \label{fig:migrated-img-and-masks}
    \end{figure}

    We split the images into training, 
        validation, 
        and test sets. 
    Figure \ref{fig:migrated-img-and-masks} illustrates the migrated images, 
        their respective binary masks, 
        and our data splits.
    
    In our experiments, 
        the images 
        and masks, 
        illustrated in figure \ref{fig:migrated-img-and-masks}, 
        are processed into patches for training and evaluating the generative 
        and semantic segmentation models.
    Moreover, 
        we use distinct strategies 
        for processing the seismic images and salt masks into patches 
        due to the different goals of the generative and semantic segmentation models. 
    In the following, 
        we describe both procedures.
    
    \subsubsection*{Patch Processing for the Generative Models}
        
        The goal of training 
            the VAE and the CNF models 
            is to generate pairs of seismic image patches 
            and their respective salt masks 
            that lie in the boundary of a salt body. 
        With this proposal, 
            the VAE and CNF models 
            are trained using only patches 
            % of the seismic images and salt masks, 
            which contain at least $10\%$ 
            and a maximum of $90\%$ of salt.
        
        Therefore, 
            the seismic images train and validation, 
            partitions are processed to create an overlapping grid 
            of $64 \times 64$ images and masks 
            such that adjacent patches in the seismic image have $90\%$ overlap area. 
        After it, 
            patches that contain at least $10\%$ 
            and a maximum of $90\%$ 
            of salt are filtered and used as training examples. 
        At last, 
            the resulting data is augmented by horizontal flipping.

\begin{figure}[ht]
            \centering
            \includegraphics[width=5cm]
            {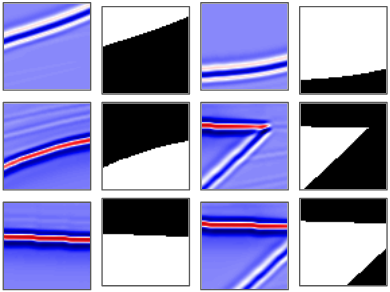}
            \caption{
            Examples of the migrated image patches and their respective salt masks. 
            In the image, 
                the second and fourth columns represent salt masks in which black regions represent non-salt areas. 
            The first and third columns represent the respective seismic representation patches.}
            \label{fig:image-size-dist}
        \end{figure}

       The resulting dataset comprises $24.872$ pairs 
        of patches and binary masks 
        containing different segments of a salt body boundary. 
        Figure \ref{fig:image-size-dist} shows examples 
            of migrated image patches and their respective salt masks.
            
    \subsubsection*{Patch Processing for the Semantic Segmentation Models} 
        
        The semantic segmentation of salt bodies task consists of predicting binary salt masks that depict salt regions from the inputted seismic image patches. 
        With the purpose 
            of training and evaluating models for this task, 
            % for the semantic segmentation of salt bodies, 
            the dataset training, 
            validation, 
            and test sets 
            are processed to create an overlapping grid of $64 \times 64$ patches. 
        The overlapping grid is created 
            such that adjacent patches in the image partition have $10\%$ overlap,
            and there are no overlapping partitions. 
        At last, 
            the data is augmented by horizontal flipping each resulting sample pair.
        
        The resulting dataset comprises $6,000$ sample pairs, 
            distributed as follows: 
            $2.450$ samples for the training set; 
            $1.434$ samples for the validation set; 
            and $2.042$ samples for the test set.
        
        Finally, 
            it is important to note that all image patches are used to create the dataset, 
            unlike the previous procedure 
            where only the patches on a salt body border are taken.

\section{Experiments} \label{sec:6-experiments}
    
    In this section, 
        we present our experiments and the achieved results. 
    The IoU (Intersection Over Union) metric at a $0.5$ threshold evaluates the model’s performance, 
        such as proposed in the PASCAL VOC challenge \cite{Everingham2009ThePV}. 
    % Thus, 
    %     the IoU metric is calculated by comparing the predicted masks against the ground truth,
    %     based on the number of true positives (TP), 
    %     false negatives (FN), 
    %     and false positives (FP) predictions. 
    % The final score is calculated as:
    % \begin{equation}
    %      \label{eq:IoU}
    %      IoU = \frac{TP}{TP + FP + FN}
    % \end{equation}
   
%   We present our experiments in the following.
   
\subsection{Augmentations Effects} \label{subsec:6.1-augmentations-effects}
    
    This experiment aims to determine how many generated pairs 
        of salt masks and image patches 
        should be used to augment the dataset 
        while training models for the semantic segmentation of salt bodies. 
    Thus, 
        we train several U-net models \cite{Ronneberger2015UNetCN} exploring different values for the augmentation size hyper-parameter, 
        which controls how many generated sample pairs are used as data augmentation. 
    The data augmentation samples are generated by 
        the method and models described in \ref{sec:4-model}.
    
    In summary, 
        the experiment consists of evaluating the IoU metric on the dataset's 
        training 
        and validation sets 
        for models trained with different values for the augmentation size, 
        from $0$ to $700$. 
    We train ten U-net models at each iteration, 
        with varying augmentation samples 
        -- i.e. pairs of seismic image patches and salt masks. 
    Subsequently, 
        the IoU metric is evaluated in training set for each model. 
    Then the model with the second-best training metric is selected, 
        and the other nine models are discarded. 
    The iteration finishes by evaluating the selected model on the validation set.
    
    % The experiment is performed as described above because it produced better 
        % and more stable results. 
    We observe that it is required to test some distinct generated augmentation sets 
        for achieving good results due to the high level of noise and bias added by our DA method.
    We found that five or fewer trials are usually sufficient 
        for finding such an augmentation set, 
        but we maintained ten shots to ensure consistency. 
    It is important to note that we select the second-best training metric 
        to avoid selecting highly over-fitted models.
    
    All models are trained with the U-net implementation\footnote{https://github.com/jakeret/unet}
        provided by \cite{akeret2017radio} for $60$ epochs, 
        using the Adam optimizer \cite{Kingma2014AdamAM} with default hyperparameters, 
        and the learning rate equals $10^{-3}$. 
    Additionally, 
        the models have four encoder and decoder blocks, 
        and the first block has $65$ filters for its convolutional layers. 
    The number of filters is doubled at each block, 
        and the whole model has $7.69M$ trainable parameters. 
    The training procedure takes 
        an average time of 10 minutes per model 
        on a Tesla P100 graphic processing unit (GPU).
    
\begin{figure}[ht]
        \centering
        \includegraphics[keepaspectratio=true,width=0.8\columnwidth]
        {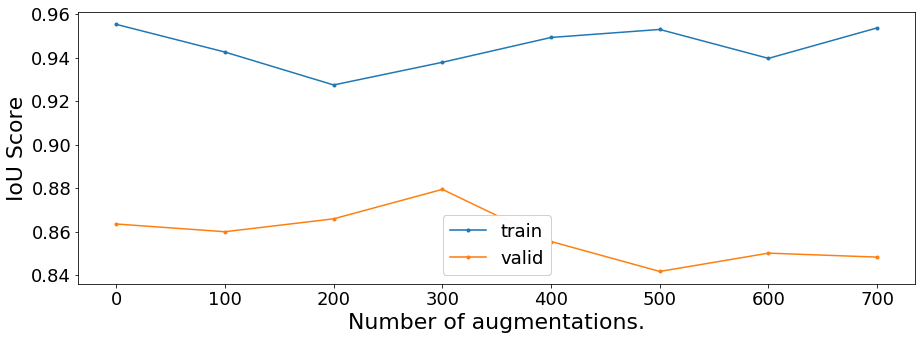}
        \caption{IoU per number of augmentations.}
        \label{fig:aug_performance_curve}
    \end{figure}

    In total, 
        we train $80$ U-Net models, 
        but we only report the metrics for the eight selected models. 
    Figure \ref{fig:aug_performance_curve} presents 
        the training and validation performance curve 
        for the selected models at different sizes of augmentations, 
        ranging from $0$ to $700$.
    
    Among the different augmentation sizes, 
        the model trained without augmentations is the one that has the highest IoU score 
        on the training set. 
    Then, 
        up to $200$ augmentations, 
        the training score decays from 
        $.96$ to $.92$, 
        which is the lowest. 
    From this point, 
        the training metric rises practically back to the initial level.
    
    On the other hand, 
        we observe an increasing validation score until we achieve $300$ augmentations, 
        which presents the highest validation score. 
    From $400$ onwards, 
        the validation IoU quickly degenerates while we observe a rising training score, 
        indicating that the model starts to overfit the augmentations from this point.
    
    Moreover, 
        observing the decay in the training metric 
        while observing the growth in the validation metric may indicate that 
        the data augmentations regularize the model during training, 
        thus avoiding overfit 
        and improving its generalization. 
    Also, 
        the highest validation metric is achieved by the model with the second-lowest training score. 
    It is observed in an inflection point, 
        from where the validation metrics start to degenerate while the training score still increases. 
    It may indicate that at $300$ augmentations, 
        the model achieves a good balance between regularization and overfitting.

\subsection{Performance Comparison} 
\label{subsec:6.2-performance-comparison}
    
    This section compares the performance of distinct models 
        when trained using 
        and not using our generated augmentation samples, 
        aiming to show that 
        our proposed method results in gains independently on the models’ architecture 
        and the number of parameters. 
    For this purpose, 
        we compare the performance of the U-Net against seven variants 
        of the DeeplabV3+ model \cite{deeplabv3plus2018}. 
    The DeeplabV3+ was chosen because its official implementation provides, 
        out-of-the-shelf, 
        several changeable backbone architectures with different depths 
        and number of parameters. 
    Three widely used architectures are explored as backbone: 
        Xception \cite{Chollet2016XceptionDL}; 
        MobileNet \cite{mobilenetv32019}; 
        Resnet \cite{He2016DeepRL};
    
    The U-net model is trained with the same configuration as explained in section \ref{subsec:6.1-augmentations-effects}, 
        and all DeeplabV3+ models are trained using the official implementation
        \footnotemark[\value{footnote}]. 
    Models with the Xception and ResNet backbones 
        have the Atrous rate, 
        output stride, 
        and decoder output stride hyper-parameters set respectively to 
        $[12,24,36]$, 
        $8$, 
        and $4$.
    Moreover, 
        all models with the MobileNet (V2 and V3) backbones, 
        the Atrous rates, and output stride hyper-parameters are left empty. 
    In models with the MobileNetV3 backbone, 
        the decoder output stride hyper-parameter is set to 8. 
    We provide more detailed hyper-parameter settings for each model on the supplementary material accompanying this paper. 
    
    Finally, 
        for comparison purposes, 
        all models are trained using the Adam Optimizer for $40K$ training iterations 
        and a mini-batch of $20$ examples per iteration 
        on an NVidia Tesla P100 graphic processing unit (GPU).
    \footnotetext{https://github.com/tensorflow/models/tree/master/research/deeplab}
    
    Table \ref{tab:1-Performance-Comparison} present the number of trainable parameters 
        and the obtained results in 
        the validation and test sets 
        for each model 
        when trained with and without the proposed DA method. 
    The augmented versions are trained with the addition of $300$ generated pairs 
        of salt masks 
        and image patches.
    
    \begin{table}[ht]
    \centering
    \caption{
        Performance comparison between models trained with and without data augmentation.
    }
    \label{tab:1-Performance-Comparison}
    \begin{tabularx}{\columnwidth}{X|lr|XX|XX}
        \toprule
        \multirow{2}{4em}{Model} &
        \multirow{2}{4em}{Backbone} &
        \multirow{2}{4em}{Trainable Parameters} &
        \multicolumn{2}{c|}{Not Augmented} & 
        \multicolumn{2}{c}{Augmented} \\
         
         &
         &
         &  $Validation$ (\%) & 
            $Test$ (\%) &   
            $Validation$ (\%) & 
            $Test$ (\%) \\
        \midrule
        
        U-net    & - & 
            % $7,697,410$ &
            $7,69M$ &
            88.66  & 
            93.03 & 
            88.57  & 
            \textbf{94.71} \\
        \midrule
        
        \multirow{9}{4em}{DeepLab} & 
            mobilenet\_v2    & 
            % $12,333,746$    &
            $2.10M$    &
            81.12  & 
            82.83 & 
            90.02  & 
            \textbf{86.83} \\
                
        % DeepLab (Mobilenet)    &
            & mobilenet\_v3\_small  &
            $.82M$    &
            66.23  & 
            71.90 & 
            83.52  & 
            \textbf{88.76} \\
        
        % DeepLab (Mobilenet)    &
            & mobilenet\_v3\_large &
            $1.91M$   &
            73.82  & 
            82.34 & 
            78.04  & 
            \textbf{89.83} \\
            
        % DeepLab (xception)    &
            & xception\_41 &
            $28.10M$   &
            77.45  & 
            75.35 & 
            88.69  & 
            \textbf{81.94} \\
        
        % DeepLab (xception)    &
            & xception\_65 &
            $41.05M$   &
            82.34  & 
            85.14 & 
            77.12  & 
            \textbf{90.11} \\
        
        % DeepLab (xception)    &
            & xception\_71 &
            $41.60M$   &
            62.99  & 
            61.86 & 
            77.42  & 
            \textbf{77.39} \\
        
        % DeepLab (resnet)    &
            & resnet\_v1\_18 &
            $12.33M$   &
            89.15  & 
            93.75 & 
            89.68  & 
            \textbf{95.17} \\
        
        % DeepLab (resnet)    &
            & resnet\_v1\_50 &
            $26.69M$   &
            87.66  & 
            88.92 & 
            85.96  & 
            \textbf{91.77} \\
        
        % DeepLab (resnet)    &
            & resnet\_v1\_101 &
            $45.68M$   &
            89.16  & 
            93.17 & 
            91.08  & 
            \textbf{95.17} \\
        \bottomrule
    \end{tabularx}
\end{table}
    
    The proposed data augmentation method yields consistent improvements for all models in the IoU metric. 
    On average, 
        we find a gain of $8.57\%$ in the test metric. 
    The best improvement is $25.10\%$, 
        found with the DeeplabV3+ using the xception\_71 backbone. 
    Also, 
        the best result is achieved with the augmented dataset by the resnet\_v1\_101 backbone.
    This model reaches $95.17\%$ in the test score, 
        which corresponds to an improvement of $2.14\%$. 
    Interestingly, 
        the models with the xception\_65 and resnet\_v1\_50 backbones 
        and the U-net model present a slight reduction in their validation metric, 
        indicating that somehow the augmentation regularized those models, 
        helping to improve their generalization on unseen data.
\subsection{Comparison against other Data Augmentation methods} 
\label{subsec:6.3-methods-comparison}
    We compare the performance of the proposed DA method 
        against seven distinct methods 
        from the Albumentation library \cite{info11020125}. 
    Thus, 
        the performance of the DeeplabV3+ with the mobilenet\_v3\_large backbone is evaluated for models trained using the following methods: 
        Elastic Transform; 
        Grid Distortion; 
        Optical Distortion; 
        CLAHE; 
        Random Brightness Contrast; 
        Random Gamma.
    % We compare the performance of models 
    %     trained using distinct DA methods against our proposed method. 
    % Thus, 
    %     the performance of the DeeplabV3+ with the mobilenet\_v3\_large\_seg backbone is evaluated
    %     for models trained using six different DA methods from the Albumentation \cite{info11020125} library namely: 
    %     Elastic Transform; 
    %     Grid Distortion; 
    %     Optical Distortion; 
    %     CLAHE; 
    %     Random Brightness Contrast; 
    %     Random Gamma. 
    Those methods were selected because they have shown noticeable results 
        for medical and image applications 
        \cite{SimardSteinkrausPlatt2003ET, Castro2018ElasticDF, HUANG2019372, agustin2020implementation}. 
    Additionally, 
        we also evaluate the models’ performance when composing our proposed DAs together with the transformations enumerated above. 
    It is important to note that 
        only the original samples are passed through the transformations. 
    Therefore, 
        our DA samples are only presented to the models without being passed through any transformation.
    
    All models are trained using the same settings described in section \ref{subsec:6.2-performance-comparison} 
        and took an average time of $1$ hour and $10$ minutes on an Nvidia Tesla P100 GPU.
    
    Table \ref{tab:2-Augmentation-Comparison} compares the obtained results, 
        in the validation and test sets, 
        of each Albumentation method 
        against our proposed data augmentation 
        and the composition of both methods.
    
    \begin{table}[ht]
    \centering
    \caption{
        Performance comparison between data augmentation methods against our proposed augmentations. 
        In the center columns, 
            we present the results achieved when each DA method stand-alone. 
        In the right columns, 
            we present the results achieved when composing each method with our proposed DA.
        % Performance comparison between data augmentation methods against our proposed augmentations.
    }
    \label{tab:2-Augmentation-Comparison}
    \begin{tabularx}{\columnwidth}{X|XX|XX}
        \toprule
        \multirow{2}{4em}{Method} &
        \multicolumn{2}{c|}{Alone} & 
        \multicolumn{2}{c}{ + Ours} \\
         
         &  \textit{Validation} (\%) & 
            \textit{Test} (\%) &   
            \textit{Validation} (\%) & 
            \textit{Test} (\%) \\
        \midrule
        
        None    & 
            73.82  & 
            82.34 & 
            -  & 
            - \\
        \midrule
        
        Ours    & 
            78.04  & 
            \textbf{89.83} & 
            -  & 
            - \\
        \midrule
        
        ElasticTransform    & 
            86.56  & 
            89.09 & 
            75.88  & 
            \textbf{90.39} \\
        
        CLAHE    & 
            81.15  & 
            89.78 & 
            78.40  & 
            86.67 \\
        
        Grid Distortion & 
            77.99  & 
            85.72 & 
            87.19  & 
            89.60 \\
                
        Random Gamma    &
            86.31  & 
            87.31 & 
            86.51  & 
            83.59 \\
        
        Random Brightness Contrast   &
            80.39  & 
            82.25 & 
            76.91  & 
            86.33 \\
        
        Optical Distortion   &
            68.06  & 
            74.90 & 
            84.26  & 
            82.21 \\

        \bottomrule
    \end{tabularx}
\end{table}
    
    Comparing the obtained results for the models trained with a single DA method 
        (the center column), 
        we observe that our method achieves the best result. 
    Moreover, 
        except for 
        the Random Brightness Contrast 
        and Optical Distortion methods, 
        all methods improve both the validation and test sets. 
    The highest test score, 
        $90.39\%$, 
        is achieved when applying the Elastic Transform 
        and our DAs together. 
    Also, 
        composing our DAs with the Optical Distortion, 
        Random Brightness Contrast, 
        and Grid Distortion methods result in better generalization than 
        when applying those methods alone. 
    Such results may indicate that our approach could complement other DA methods,
        improving the models’ performance even more. 
    On the other hand, 
        it is not all methods that improve results when applied with our DAs together. 
    The CLAHE and Random Gamma methods present better results when used alone. 
    Unfortunately, 
        we could not identify an apparent reason for these results.
    
\subsection{Performance comparison on a larger context size} 
\label{subsec:6.4-experiment-with-larger-context-size}
    
    In this experiment, 
        we compare the performance of the DeeplabV3+ using the mobilenet\_v3\_large backbone 
        when trained using 
        and not using our generated DAs for $128 \times 128$ patches 
        aiming to demonstrate that our proposal also works with larger context sizes, 
        i.e., larger patch sizes.
    % First, 
    For this purpose,
        we train the generative models, 
        with the same setup described in section \ref{sec:4-model},
        for $128 \times 128$ context size. 
    The context patches are processed to create an overlapping grid of $128 \times 128$ 
        with $90\%$ overlap between adjacent patches. 
    Additionally, 
        only the patch samples containing between $10\%$ and $90\%$ of salt are used. 
    The data is also augmented by applying horizontal flips, 
        resulting in $11.056$ samples. 
    
    The CNF model achieves a loss value of $1.81$ after six days 
        and $259,2K$ training iterations 
        on a Google Cloud TPU v2-8. 
    In its turn, 
        the VAE loss converged around $61.17$ after $14$ hours and $28.8K$ training iterations 
        on an NVidia Tesla P100 GPU.
    
    For the semantic segmentation model, 
        $128\times128$ context patches, 
        with $10\%$ overlap between adjacent patches, 
        are created, 
        and the data is also augmented by applying horizontal flips.
    % The resulting dataset contains $1,476$ samples, 
    %     being $608$ in the train,
    %     $356$ in the validation, 
    %     and $512$ in the test sets. 
    The resulting dataset contains $1,476$ samples, 
        with $608$ in the training set, 
        $356$ in the validation set, 
        and $512$ in the test set.
    The DeeplabV3+ is trained during 40k iterations three times for
        augmented and not augmented versions, 
        taking an average time of $6$ hours and $20$ minutes per model.
    % The DeeplabV3+ is trained during $40k$ iterations three times for augmented 
        % and not augmented versions. 
    Table \ref{tab:3-128-context-performance-comparison} compares the performance of the model 
        with the best validation score of each version.
    
    \begin{table}[ht]
    \centering
    \caption{
        Performance comparison of the DeeplabV3+ model with the mobilenet\_v3\_large backbone for context size of $128 \times 128$ when trained with and without data augmentation.
    }
    \label{tab:3-128-context-performance-comparison}
    \begin{tabularx}{\columnwidth}{X|XX}
        \toprule
        Version &
        $Validation$ (\%) & 
        $Test$ (\%) \\  
        \midrule
        
        Not Augmented    & 
            74.09 &
            84.85 \\

        \textbf{Augmented} & 
            \textbf{86.19}   & 
            \textbf{90.55}   \\

        \bottomrule
    \end{tabularx}
\end{table}
    
    The augmented version improved the metrics on the validation and test sets 
        from $74.09\%$ to $86.19\%$ 
        and from $84.85\%$ to $90.55\%$, 
        respectively. 
    When compared against the results for the same model architecture, 
        from experiment 6.2 (in $64 \times 64$ contexts), 
        the augmented and not augmented models outperformed their respective versions. 
    Moreover, 
        comparing the improvement percentage in the test set of the models trained for $64 \times 64$ contexts against the models trained for $128 \times 128$ contexts, 
        we observe that although the percentage reduced from $9.09\%$, 
        to $6.71\%$, 
        we still have a significant gain when training the model with our proposed DA. 
    We believe that the obtained results show the potential of the proposed DA method on larger context sizes. 
    Finally, 
        it is important to note that further investigation into context sizes has become prohibitive in this work due to the many executed experiments and the high computational cost required for this.
    
\section{Conclusion} \label{sec:7-conclusion}
    
    This work proposes a Data Augmentation (DA) method 
        for the semantic segmentation of salt bodies in a seismic image dataset 
        built from two public available synthetic salt models.
    
    The performance of distinct state-of-the-art semantic segmentation models is compared 
    when using and not using the proposed DAs during the training procedure. 
    The proposed method yields consistent improvements independently from 
        the model architecture 
        or their number of trainable parameters. 
    Moreover, 
        our DA method achieves an average improvement of $8.57\%$ in the IoU metric.
    % The best improvement is $25.10\%$, 
        % found with the DeeplabV3+ using the xception\_71 backbone. 
    The best improvement, 
        of $25.10\%$, 
        is achieved by the DeeplabV3+ using the the xception\_71 backbone. 
    In it turns, 
        the best result is achieved when augmenting data for the DeeplabV3+ with the resnet\_v1\_101 backbone. 
    This model reaches $95.17\%$ in the test score, 
        which corresponds to an improvement of $2.14\%$. 
    
    We also show that our proposal outperforms seven selected DA methods when they are applied stand-alone. 
    Furthermore, 
        by applying our data augmentations in composition with the ElastTransform augmentation method,  
        DeeplabV3+ using the large version of the MobileNetV3 backbone achieves a $90.39\%$ score, 
        which is the highest for this architecture.
    At last, 
        the same methodology for $128 \times 128$ context patches is applied, 
        achieving results comparable to the obtained for smaller contexts, 
        showing that our proposal is adaptable for larger contexts.

    Future work could include 
        testing the proposed method in other datasets, 
        such as the one used in the TGS Salt Identification Challenge;
        application in other seismic tasks like the seismic facies segmentation \cite{Zhao2015ACO, Wrona2018SEISMICFA, Dramsch2018DeeplearningSF};
        adapting the methodology for natural images; 
        and exploring the use of other generative models such as the Conditional GAN \cite{Mirza2014ConditionalGA}.
    
    Finally, 
        our experimental results show that the proposed DA method yields consistent improvements for different models and context sizes, 
        helping develop models for the semantic segmentation of salt bodies. 
\section*{Acknowledgment}
This material is based upon work supported by PETROBRAS, CENPES.
Also, this material is based upon work supported by Air Force Office Scientific Research under award number FA9550-19-1-0020.

\bibliographystyle{unsrt}  
\bibliography{references}

\end{document}